%% file: main.tex
\documentclass[twoside,11pt, usenames,dvipsnames]{article}

\usepackage[abbrvbib, preprint]{jmlr2e} % for preprint
\usepackage{todonotes}
\usepackage{amsfonts}
\usepackage{pythonhighlight}
\usepackage[flushleft]{threeparttable}
\usepackage{booktabs}
\usepackage{algorithm}
\usepackage{algpseudocode}
\usepackage{multirow}
\usepackage{rotating}
\usepackage{pdflscape}
\usepackage{enumitem}
\usepackage{amsmath}

\usepackage{amssymb}% http://ctan.org/pkg/amssymb
\usepackage{pifont}% http://ctan.org/pkg/pifont
\usepackage[makeroom]{cancel}
\newcommand{\cmark}{{\color{ForestGreen} \ding{51}}}%
\newcommand{\xmark}{{\color{red}\ding{55}}}%
\newcommand{\pmark}{{\color{blue}\bcancel{\ding{51}}}}%

\usepackage{lastpage}

\firstpageno{1}

\begin{document}

\title{\texttt{TorchSurv}: A Lightweight Package for Deep Survival Analysis}

\author{\name Mélodie Monod \email melodie.monod@novartis.com \\
       \addr Novartis Pharma AG, Switzerland
       \AND
       \name Peter Krusche \email peter.krusche@novartis.com \\
       \addr Novartis Pharma AG, Switzerland
       \AND
       \name Qian Cao \email qian.cao@fda.hhs.gov \\
       \addr Center for Devices and Radiological Health, Food and Drug Administration, MD, USA
       \AND
       \name Berkman Sahiner \email berkman.sahiner@fda.hhs.gov \\
       \addr Center for Devices and Radiological Health, Food and Drug Administration, MD, USA
       \AND
       \name Nicholas Petrick \email nicholas.petrick@fda.hhs.gov \\
       \addr Center for Devices and Radiological Health, Food and Drug Administration, MD, USA
       \AND
       \name David Ohlssen \email david.ohlssen@novartis.com \\
       \addr Novartis Pharmaceuticals Corporation, NJ, USA
       \AND
       \name Thibaud Coroller\textsuperscript{@} \email thibaud.coroller@novartis.com \\
       \addr Novartis Pharmaceuticals Corporation, NJ, USA\\
       \textsuperscript{@}Corresponding author}
\editor{My editor}

\maketitle

%%==================================%%

%% ABSTRACT %%

%%==================================%%

\begin{abstract}
\texttt{TorchSurv}\footnote{The source code of \texttt{TorchSurv} can be found at~\url{https://github.com/Novartis/torchsurv}.}\footnote{The Python Package Index of \texttt{TorchSurv} can be found at~\url{https://pypi.org/project/torchsurv}.}\footnote{The latest documentation of \texttt{TorchSurv} can be found at~\url{https://opensource.nibr.com/torchsurv/}.} is a Python package that serves as a companion tool to perform deep survival modeling within the \texttt{PyTorch} environment. Unlike existing libraries that impose specific parametric forms, \texttt{TorchSurv} enables the use of custom \texttt{PyTorch}-based deep survival models. 
% It offers comprehensive functionalities for computing log-likelihoods of common survival models and evaluating their predictive performance.
With its lightweight design, minimal input requirements, full \texttt{PyTorch} backend, and freedom from restrictive survival model parameterizations, \texttt{TorchSurv} facilitates efficient deep survival model implementation and is particularly beneficial for high-dimensional and complex input data scenarios.
% \texttt{torchsurv} distinguishes itself by providing a simple and accessible programming interface with \texttt{PyTorch} backend.

\end{abstract}


%%==================================%%

%% MAIN TEXT %%

%%==================================%%

%
% Introduction
\section{Introduction} \label{sec:introduction}
\input{introduction}

% Related work
\section{Related Work} \label{sec:related}

\input{related_work}
\section{Core Functions} \label{sec:functions}
\input{functions}

%
% Illustrative examples
% \section{Illustrative examples} \label{sec:examples}
% \input{examples}

%
% Conclusion
\section{Conclusion} \label{sec:conclusion}
\input{conclusion}

%
%  Acknowledgments
%\section*{Acknowledgments}
% This was was supported in part by......

%  COI
\section{Disclosures}

MM, PK, DO and TC are employees and stockholders of Novartis, a global pharmaceutical company.

%%==================================%%

%% POSSIBLE APPENDIX %%

%%==================================%%

\newpage
\begin{landscape}
\appendix
\section{Survival Analysis Libraries in R} \label{supp:survival_library_R}
\input{table_R}
\end{landscape}

%%==================================%%

%% Bibliography %%

%%==================================%%

\bibliography{references}  
\end{document}

%% file: introduction.tex
Survival analysis plays a crucial role in various domains, such as medicine and engineering. Deep learning offers promising avenues for building complex survival models. However, existing libraries often restrict users to predefined parameter forms and limit seamless integration with \texttt{PyTorch} \citep{paszke2019pytorch}. We introduce \texttt{TorchSurv} (available on GitHub and PyPI), a toolkit to empower researchers in building and evaluating deep survival models within the \texttt{PyTorch} framework. 
\texttt{TorchSurv} provides a user-friendly workflow for defining a survival model with parameters specified by a \texttt{PyTorch}-based (deep) neural network. Training is facilitated by leveraging one of \texttt{TorchSurv}'s built-in survival model loss functions. Upon completion of model training, \texttt{TorchSurv} provides metrics for evaluating the survival model's predictive performance.
\texttt{TorchSurv} has undergone rigorous testing on open source data and synthetically generated survival data that include edge cases. The package is comprehensively documented and contains illustrative examples. The latest documentation of \texttt{TorchSurv} can be found at~\url{https://opensource.nibr.com/torchsurv/}.

At the core of \texttt{TorchSurv} lies its \texttt{PyTorch}-based calculation of log-likelihoods for prominent survival models, including the Cox proportional hazards model~\citep{Cox1972} and the Weibull Accelerated Time Failure (AFT) model~\citep{Carroll2003}.
In survival analysis, each observation is associated with survival data denoted by $y$ (comprising the event indicator and the time-to-event or censoring) and covariates denoted by $x$. A survival model that is able to capture the complexity of the survival data $y$, is parametrized by parameters denoted by $\theta$. For instance, in the Cox proportional hazards model, the survival model parameters $\theta$ are the relative hazards. Within the \texttt{TorchSurv} framework, a \texttt{PyTorch}-based neural network is defined to act as a flexible function that takes the covariates $x$ as input and outputs the survival model parameters $\theta$. 
\texttt{TorchSurv}'s log-likelihoods are calculated from the survival data $y$ and the survival model parameters $\theta$. Estimation of the parameters $\theta$ is achieved via maximum likelihood estimation facilitated by backpropagation. To allow for automatic gradient computation and enable maximum likelihood estimation, the log-likelihood computations are fully implemented in \texttt{PyTorch}. 
The automatic gradient calculation is crucial as it allows for efficient optimization of complex models by automatically computing the gradients of the loss function with respect to the survival model parameters $\theta$. To mitigate numerical instability and ensure stable training, all computations are conducted on the logarithmic scale.
%To exemplify the practical benefits of \texttt{TorchSurv}, consider the scenario where a user seeks to fit a Cox proportional hazards Deep Neural Network in \texttt{PyTorch}. This model assumes the hazard function to be the product of a baseline hazard, dependent solely on time, and individual-specific relative hazards, independent of time but dependent on a set of covariates. The function mapping the set of covariates to relative hazards is represented by a (deep) neural network. The negative of the Cox proportional hazards's partial log-likelihood is a natural choice for the loss function. However, no existing package provides a lightweight version of the Cox partial log-likelihood in \texttt{PyTorch}. In the current context, users have two options to fit this model. Either they adopt the deep neural network's architecture from an existing package, such as deepsurv~\citep{katzman2018deepsurv}, or they must code the partial log-likelihood themselves. 

Additionally, \texttt{TorchSurv} offers evaluation metrics to characterize the predictive performance of survival models. These evaluation metrics include the time-dependent Area Under the cure (AUC) under the Receiver operating characteristic curve (ROC), the Concordance index (C-index) and the Brier Score. The evaluation metric's functionalities include the point estimate, confidence interval, hypothesis test to determine whether the metric is better than that of a random predictor and hypothesis test to compare two metrics obtained with different models.

%To reiterate the example above, after fitting the model, the metrics provided by \texttt{TorchSurv} enable users to evaluate the predictive performance of their model, without the need for extraneous data conversion (e,g., from \texttt{PyTorch} tensors to \texttt{NumPy} arrays) or the imposition of a specific output format. 

% The paper is structured as follows. Section~\ref{sec:functions} details the core function of the \texttt{TorchSurv} package. Section~\ref{sec:examples} presents illustrative examples on open-source data. Lastly, Section~\ref{sec:conclusion} closes with a discussion.

%% file: related_work.tex
Table~\ref{tab:bibliography} compares the functionalities of \texttt{TorchSurv} with those of 
\texttt{auton-survival}~\citep{nagpal2022auton}, 
\texttt{pycox}~\citep{Kvamme2019pycox}, 
\texttt{torchlife}~\citep{torchlifeAbeywardana},
\texttt{scikit-surv-} \texttt{ival}~\citep{polsterl2020scikit}
\texttt{lifelines}~\citep{davidson2019lifelines}, and 
\texttt{deepsurv}~\citep{katzman2018deepsurv}.
While several libraries offer survival modelling functionalities, as shown in Table~\ref{tab:bibliography}, no existing library provides the flexibility to use a custom \texttt{PyTorch}-based neural network to define the survival model parameters $\theta$ given a set of covariates $x$. In existing libraries, users are limited to specific forms to define $\theta$ (e.g., linear function of covariates) and the log-likelihood functions available cannot be leveraged because they do not allow for seamless integration with \texttt{PyTorch}.
Specifically, the limitations on the log-likelihood functions include protected functions, specialized input requirements (format or class type), and reliance on external libraries like \texttt{NumPy} or \texttt{Pandas}. Dependence on external libraries hinders automatic gradient calculation within \texttt{PyTorch}. Additionally, the implementation of likelihood functions instead of log-likelihood functions, as done by some packages, introduces potential numerical instability.
With respect to the evaluation metrics, \texttt{scikit-survival} stands out as a comprehensive library. However, it lacks certain desirable features, including confidence intervals and comparison of the evaluation metric between two different models, and it is implemented with \texttt{NumPy}. 
Our package, \texttt{TorchSurv}, is specifically designed for use in Python, but we also provide a comparative analysis of its functionalities with popular \texttt{R} packages for survival analysis in Appendix~\ref{supp:survival_library_R}. \texttt{R} packages do not make log-likelihood functions readily accessible and restrict users to specific forms to define  $\theta$. However, \texttt{R} has extensive libraries for evaluation metrics, such as the \texttt{RiskRegression} library~\citep{riskRegressionpackage}. \texttt{TorchSurv} offers a comparable range of evaluation metrics, ensuring comprehensive model evaluation regardless of the chosen programming environment. 

\input{table_python}

%% file: table_python.tex
\begin{table}[t!]
  \centering
  \begin{threeparttable}
  \resizebox{1\textwidth}{!}{%
  \begin{tabular}{l l l l l l l l}
  \toprule
     & \texttt{TorchSurv}  & \texttt{auton\_survival}\tnote{$1$} & \texttt{pycox}\tnote{$2$} & \texttt{torchlife}\tnote{$3$} & \texttt{scitkit-survival}\tnote{$4$} & \texttt{lifelines}\tnote{$5$} & \texttt{deepsurv}\tnote{$6$} \\
    \midrule
    \textbf{PyTorch} & \cmark & \cmark & \cmark & \cmark & \xmark & \xmark & \xmark \\ 
    \midrule
    \multicolumn{8}{l}{\textbf{Standalone loss functions}}   \\
    \quad Weibull loss function &  \cmark  & \xmark & \xmark & \cmark &  \xmark & \cmark & \xmark \\
    \quad Cox loss function  &  \cmark  &  \cmark &  \cmark & \cmark&  \cmark &  \cmark & \xmark \\
     \quad  \quad Handle ties in event time  &  \cmark & \xmark & \xmark & \xmark&  \pmark &  \cmark & \xmark \\
     \quad Logarithm scale &  \cmark&  \cmark&  \cmark&  \xmark&  \cmark&  \cmark&  \xmark\\
    \toprule
    \multicolumn{8}{l}{\textbf{Standalone evaluation metrics}} \\
    \quad Concordance index  & \cmark & \xmark& \pmark& \xmark& \cmark& \cmark& \xmark \\ 
    \quad AUC  & \cmark & \xmark& \xmark& \xmark& \cmark& \xmark& \xmark \\ 
    \quad Brier-Score  & \cmark & \xmark& \cmark& \xmark& \cmark& \xmark& \xmark \\ 
    \quad  Time-dependent risk score   & \cmark & \xmark   & \pmark & \xmark & \pmark & \xmark & \xmark\\
    \quad Subject-specific weights      & \cmark & \xmark   & \pmark & \xmark& \pmark & \xmark & \xmark\\
    \quad Confidence interval      & \cmark & \xmark   & \xmark & \xmark& \xmark & \xmark & \xmark\\
    \quad Compare two metrics & \cmark & \xmark   & \xmark & \xmark& \xmark & \xmark & \xmark\\
    \midrule
    \textbf{Momentum} & \cmark & \xmark   & \xmark & \xmark& \xmark & \xmark & \xmark\\
  \bottomrule
  \end{tabular}}
  \begin{tablenotes}
  \item[$1$] \citep{nagpal2022auton}, $^{2}$ \citep{Kvamme2019pycox}, $^{3}$ \citep{torchlifeAbeywardana}, $^{4}$ \citep{polsterl2020scikit},   
  \item[$^{5}$] \citep{davidson2019lifelines}, $^{6}$ \citep{katzman2018deepsurv}. \cmark $\,$ indicates a fully supported feature, 
  \item[] \xmark $\,$ indicates an unsupported feature, \pmark $\,$  indicates a partially supported feature. 
  \end{tablenotes}
  \end{threeparttable}
    \caption{\textbf{Survival analysis libraries in Python.} For computing the concordance index, \texttt{pycox} requires the use of the estimated survival function as the risk score and does not support other types of time-dependent risk scores. \texttt{scikit-survival} does not support time-dependent risk scores in both the concordance index and AUC computation. Additionally, both \texttt{pycox} and \texttt{scikit-survival} impose the use of inverse probability of censoring weighting (IPCW) for subject-specific weights. \texttt{scikit-survival} only offers the Breslow approximation of the Cox partial log-likelihood in case of ties in the event time, while it lacks the Efron approximation.} 
    
  \label{tab:bibliography}
\end{table}

%% file: functions.tex
%
%
% OVERVIEW
%
%

\subsection{Overview}
The \texttt{TorchSurv} library provides loss functions to estimate survival model's parameters using maximum likelihood estimation through backpropagation. In addition, the library offers a set of evaluation metrics to characterize the predictive performance of survival models. Below is an overview of the workflow for model inference and evaluation with \texttt{TorchSurv}:

\begin{enumerate}[itemsep=0cm, parsep=0cm, topsep=0.2cm, partopsep=0cm]
    \item Initialize a \texttt{PyTorch}-based neural network that defines the function from the covariates to the survival model's parameters.
    \item Initiate training: For each epoch on the training set,
    \begin{itemize}[label=$\circ$, itemsep=0cm, parsep=0cm, topsep=0.1cm, partopsep=0cm]
        \item Draw survival data $y^{\text{train}}$ (i.e., event indicator and time-to-event or censoring) and covariates $x^{\text{train}}$ from the training set.
        \item Obtain parameters $\theta^{\text{train}}$ based on drawn covariates $x^{\text{train}}$ using \texttt{PyTorch}-based neural network.
        \item Calculate the loss given survival data $y^{\text{train}}$ and parameters $\theta^{\text{train}}$ using \texttt{TorchSurv}'s loss function.
        \item Utilize backpropagation to update parameters $\theta^{\text{train}}$.
    \end{itemize}
    \item Obtain parameters $\theta^{\text{test}}$ based on covariates from the test set $x^{\text{test}}$ using the trained \texttt{PyTorch}-based neural network.
    \item Evaluate the predictive performance of the model using \texttt{TorchSurv}'s evaluation metric functions (e.g., C-index) given parameters $\theta^{\text{test}}$ and survival data from the test set $y^{\text{test}}$.
\end{enumerate}
The outputs of both the log-likelihood functions and the evaluation metrics functions have undergone thorough comparison with benchmarks generated with Python packages (Table~\ref{tab:bibliography}) and R packages (Appendix~\ref{supp:survival_library_R}) on open-source data and synthetic data. High agreement between the outputs is consistently observed, providing users with confidence in the accuracy and reliability of \texttt{TorchSurv}'s functionalities. The comparison is summarized on the package's website at this link:~\url{https://opensource.nibr.com/torchsurv/benchmarks.html}.

%
%
% LOSS FUNCTIONS
%
%

\subsection{Loss Functions}

\paragraph{Cox loss function.} The Cox loss function is defined as the negative of the Cox proportional hazards model's partial log-likelihood~\citep{Cox1972}. The function requires the subject-specific log relative hazards and the survival data (i.e., event indicator and time-to-event or censoring). The log relative hazards should be obtained from a \texttt{PyTorch}-based model pre-specified by the user. In case of ties in the event times, the user can choose between the Breslow~\citep{Breslow1975} and the Efron method~\citep{Efron1977} to approximate the Cox partial log likelihood. We illustrate the use of the Cox loss function for a pseudo training loop in the code snippet below.

\begin{python}
from torchsurv.loss import cox
my_model = MyPyTorchModel() # PyTorch model for log hazards (1 output)
for data in dataloader:
    x, event, time = data
    log_hzs = my_model(x)  # torch.Size([64, 1]), if batch size is 64 
    loss = cox.neg_partial_log_likelihood(log_hzs, event, time)  
    loss.backward()  # native torch backend
\end{python}

\paragraph{Weibull loss function.} The Weibull loss function is defined as the negative of the Weibull AFT's log-likelihood~\citep{Carroll2003}. The function requires the subject-specific log parameters of the Weibull distribution (i.e., the log scale and the log shape) and the survival data. The log parameters of the Weibull distribution should be obtained from a \texttt{PyTorch}-based model pre-specified by the user. We illustrate the use of the Weibull loss function for a pseudo training loop in the code snippet below.

\begin{python}
from torchsurv.loss import weibull
my_model = MyPyTorchModel() # PyTorch model for log parameters (2 outputs)
for data in dataloader:
    x, event, time = data
    log_params = my_model(x)  # torch.Size([64, 2]), if batch size is 64 
    loss = weibull.neg_log_likelihood(log_params, event, time)  
    loss.backward()  # native torch backend
\end{python}

\paragraph{Momentum.} When training a model with a large file, the batch size is greatly limited by computational resources. This impacts the stability of model optimization, especially when rank-based loss is used. Inspired from MoCO \citep{he2020momentum}, we implemented a momentum loss that decouples batch size from survival loss, increasing the effective batch size and allowing robust train of a model, even when using a very limited batch size (e.g., $batch_{size} \leq 16$). We illustrate the use of momentum for a pseudo training loop in the code
snippet below.

\begin{python}
from torchsurv.loss import Momentum
my_model = MyPyTorchModel() # PyTorch model for log hazards (1 output)
my_loss = cox.neg_partial_log_likelihood  # any torchsurv loss
momentum = Momentum(backbone=my_model, loss=my_loss) 

for data in dataloader:
    x, event, time = data
    loss = model_momentum(x, event, time)  # torch.Size([64, 1])
    loss.backward()  # native torch backend

# Inference is computed with target network (k)
log_hzs = model_momentum.infer(x)  # torch.Size([64, 1])
\end{python}

%
%
% EVALUATION METRICS
%
%

\subsection{Evaluation Metrics Functions}

The \texttt{TorchSurv} package offers a comprehensive set of metrics to evaluate the predictive performance of survival models, including the AUC, C-index, and Brier score. The inputs of the evaluation metrics functions are the individual risk score estimated on the test set and the survival data on the test set. The risk score measures the risk (or a proxy thereof) that a subject has an event.  We provide definitions for each metric and demonstrate their use through illustrative code snippets. 

\paragraph{AUC.} The AUC measures the discriminatory capacity of a model at a given time $t$, i.e., the model’s ability to provide a reliable ranking of times-to-event based on estimated individual risk scores~\citep{Heagerty2005,Uno2007,Blanche2013}. 
%The AUC is recommended for evaluating time-dependent predictions (e.g., 10-year mortality) instead of the C-index, as the AUC is proper in this context, while the concordance index is not~\citep{Blanche2018}.

\begin{python}
from torchsurv.metrics import Auc
auc = Auc()
auc(log_hzs, event, time)  # AUC at each time 
auc(log_hzs, event, time, new_time=torch.tensor(10.))  # AUC at time 10
\end{python}

\paragraph{C-index.} The C-index is a generalization of the AUC that represents the assessment of the discriminatory capacity of the model over time~\citep{Harrell1996, Uno_2011}.

\begin{python}
from torchsurv.metrics import ConcordanceIndex
cindex = ConcordanceIndex()
cindex(log_hzs, event, time)  # c-index
\end{python}

\paragraph{Brier score.} The Brier score evaluates the accuracy of a model at a given time $t$. It represents the average squared distance between the observed survival status and the predicted survival probability~\citep{Graf_1999}. The Brier score cannot be obtained for the Cox model because the survival function is not available, but it can be obtained for the Weibull model.

\begin{python}
from torchsurv.metrics import Brier
surv = survival_function(log_params, time) 
brier = Brier()
brier(surv, event, time)  # Brier score at each time
brier.integral() # integrated brier score
\end{python} 

In \texttt{TorchSurv}, the evaluation metrics can be obtained for time-dependent and time-independent risk scores (e.g., for proportional and non-proportional hazards). Additionally subjects can be optionally weighted (e.g., by the inverse probability of censoring weighting (IPCW)).
Lastly, functionalities including the confidence interval, one-sample hypothesis test to determine whether the metric is better than that of a random predictor, and two-sample hypothesis test to compare two evaluation metrics between  different models are implemented. For the hypothesis tests, the significance level, typically referred to as $\alpha$, can be modified as needed. The following code snippet exemplifies the aforementioned functionalities for the C-index.

\begin{python}
cindex.confidence_interval()  # CI, default alpha = .05
cindex.p_value(alternative='greater')  # pvalue, H0: c = 0.5, HA: c > 0.5
cindex.compare(cindex_other)  # pvalue, H0: c1 = c2, HA: c1 > c2
\end{python}

%% file: conclusion.tex
This paper introduces \texttt{TorchSurv}, a Python package for deep survival modeling in  \texttt{PyTorch}. Unlike existing libraries, \texttt{TorchSurv} allows users to define custom \texttt{PyTorch}-based deep survival models, offering crucial flexibility for complex data. \texttt{TorchSurv} provides extensive functionalities for computing log-likelihoods and evaluating the predictive performance of survival models efficiently. In summary, \texttt{TorchSurv} is a flexible toolkit for researchers and practitioners, enhancing the capabilities of deep survival modeling in the \texttt{PyTorch} framework.

%% file: table_R.tex
\begin{table}[h!]

  \centering
  \begin{threeparttable}
  \resizebox{1.36\textwidth}{!}{%
  \begin{tabular}{l l l l l l l}
  \cmidrule[\heavyrulewidth]{1-7}
  \cmidrule[\heavyrulewidth]{1-7}
     & \texttt{TorchSurv}  & \texttt{survival} & \texttt{survAUC} & \texttt{timeROC} & \texttt{RisksetROC} & \texttt{survcomp}  \\
      &    & \citep{survivalpackage} & \citep{survAUCpackage} & \citep{timeROCpackage} & \citep{risksetROCpackage} & \citep{survcomppackage}  \\
   \cmidrule[\heavyrulewidth]{1-7}
    \multicolumn{5}{l}{\textbf{Standalone evaluation metrics}} \\
    \quad Concordance index  & \cmark & \pmark& \cmark& \xmark& \pmark& \cmark\\ 
    \quad AUC  & \cmark & \xmark& \cmark& \cmark& \pmark& \pmark \\ 
    \quad Brier-Score  & \cmark & \xmark& \cmark& \xmark& \xmark& \xmark\\ 
    \quad  Time-dependent risk score   & \cmark & \cmark   & \xmark & \xmark & \xmark & \xmark\\
    \quad Subject-specific weights & \cmark & \cmark   & \cmark & \pmark & \xmark & \xmark\\
    \quad Confidence interval & \cmark & \xmark   & \xmark & \cmark & \xmark & \cmark\\
    \quad Compare two metrics & \cmark & \xmark   & \xmark & \cmark & \xmark & \cmark\\
    \quad Competing risks setting & \xmark & \xmark   & \xmark & \cmark & \xmark & \xmark\\
  \cmidrule[\heavyrulewidth]{1-7}
  \cmidrule[\heavyrulewidth]{1-7}
  \\
  \cmidrule[\heavyrulewidth]{1-6}
  \cmidrule[\heavyrulewidth]{1-6}
     & \texttt{TorchSurv}  & \texttt{SurvivalROC} & \texttt{riskRegression} & \texttt{SurvMetrics} & \texttt{pec}  \\
      &   & \citep{survivalROCpackage} & \citep{riskRegressionpackage} & \citep{SurvMetricspackage} & \citep{pecpackage} \\
    \cmidrule[\heavyrulewidth]{1-6}
    \multicolumn{4}{l}{\textbf{Standalone evaluation metrics}} \\
    \quad Concordance index  & \cmark & \xmark& \xmark & \pmark& \pmark\\ 
    \quad AUC  & \cmark  & \pmark& \pmark& \xmark& \xmark\\ 
    \quad Brier-Score  & \cmark & \xmark& \pmark & \pmark & \pmark\\ 
    \quad  Time-dependent risk score   & \cmark & \xmark& \pmark& \cmark& \cmark\\
    \quad Subject-specific weights      & \cmark & \xmark& \pmark& \xmark& \pmark\\
    \quad Confidence interval      & \cmark & \xmark& \cmark& \xmark& \cmark\\
    \quad Compare two metrics & \cmark & \xmark& \cmark& \xmark& \xmark\\
    \quad Competing risks setting & \xmark & \xmark& \cmark& \cmark& \cmark\\
  \cmidrule[\heavyrulewidth]{1-6}
  \cmidrule[\heavyrulewidth]{1-6}
  \end{tabular}}
  \begin{tablenotes}
  \item[] \cmark $\,$ indicates a fully supported feature, \xmark $\,$ indicates an unsupported feature, \pmark $\,$  indicates a partially supported feature. 
  \end{tablenotes}
  \end{threeparttable}
    \caption{\textbf{Survival analysis libraries in R.} For obtaining the evaluation metrics, packages \texttt{survival}, \texttt{riskRegression}, \texttt{SurvMetrics} and \texttt{pec} require the fitted model object as input (a specific object format) and \texttt{RisksetROC} imposes a smoothing method. Packages \texttt{timeROC}, \texttt{riskRegression} and \texttt{pec} force the user to choose a form for subject-specific weights (e.g., inverse probability of censoring weighting (IPCW)). 
    %For obtaining the concordance index, \texttt{SurvMetrics} does not allow subject-specific weight. 
    Packages \texttt{survcomp} and \texttt{SurvivalROC} do not implement the general AUC but the censoring-adjusted AUC estimator proposed by~\citet{Heagerty2000}. } 
    
  \label{tab:library_R}
\end{table}

%% file: main.bbl
\begin{thebibliography}{28}
\providecommand{\natexlab}[1]{#1}
\providecommand{\url}[1]{\texttt{#1}}
\expandafter\ifx\csname urlstyle\endcsname\relax
  \providecommand{\doi}[1]{doi: #1}\else
  \providecommand{\doi}{doi: \begingroup \urlstyle{rm}\Url}\fi

\bibitem[Abeywardana(2021)]{torchlifeAbeywardana}
S.~Abeywardana.
\newblock \emph{torchlife: Survival Analysis using pytorch}, 2021.
\newblock URL \url{https://sachinruk.github.io/torchlife//index.html}.

\bibitem[Blanche(2019)]{timeROCpackage}
P.~Blanche.
\newblock \emph{Time-Dependent ROC Curve and AUC for Censored Survival Data}, 2019.
\newblock URL \url{https://CRAN.R-project.org/package=timeROC}.
\newblock R package version 0.4.

\bibitem[Blanche et~al.(2013)Blanche, Dartigues, and Jacqmin‐Gadda]{Blanche2013}
P.~Blanche, J.~Dartigues, and H.~Jacqmin‐Gadda.
\newblock Review and comparison of roc curve estimators for a time‐dependent outcome with marker‐dependent censoring.
\newblock \emph{Biometrical Journal}, 55\penalty0 (5):\penalty0 687–704, June 2013.
\newblock ISSN 1521-4036.
\newblock \doi{10.1002/bimj.201200045}.
\newblock URL \url{http://dx.doi.org/10.1002/bimj.201200045}.

\bibitem[Breslow(1975)]{Breslow1975}
N.~E. Breslow.
\newblock Analysis of survival data under the proportional hazards model.
\newblock \emph{International Statistical Review / Revue Internationale de Statistique}, 43\penalty0 (1):\penalty0 45, Apr. 1975.
\newblock ISSN 0306-7734.
\newblock \doi{10.2307/1402659}.
\newblock URL \url{http://dx.doi.org/10.2307/1402659}.

\bibitem[Carroll(2003)]{Carroll2003}
K.~J. Carroll.
\newblock On the use and utility of the weibull model in the analysis of survival data.
\newblock \emph{Controlled Clinical Trials}, 24\penalty0 (6):\penalty0 682–701, Dec. 2003.
\newblock ISSN 0197-2456.
\newblock \doi{10.1016/s0197-2456(03)00072-2}.
\newblock URL \url{http://dx.doi.org/10.1016/S0197-2456(03)00072-2}.

\bibitem[Cox(1972)]{Cox1972}
D.~R. Cox.
\newblock Regression models and life‐tables.
\newblock \emph{Journal of the Royal Statistical Society: Series B (Methodological)}, 34\penalty0 (2):\penalty0 187–202, Jan. 1972.
\newblock ISSN 2517-6161.

\bibitem[Davidson-Pilon(2019)]{davidson2019lifelines}
C.~Davidson-Pilon.
\newblock lifelines: survival analysis in python.
\newblock \emph{Journal of Open Source Software}, 4\penalty0 (40):\penalty0 1317, 2019.

\bibitem[Efron(1977)]{Efron1977}
B.~Efron.
\newblock The efficiency of cox’s likelihood function for censored data.
\newblock \emph{Journal of the American Statistical Association}, 72\penalty0 (359):\penalty0 557–565, Sept. 1977.
\newblock ISSN 1537-274X.
\newblock \doi{10.1080/01621459.1977.10480613}.
\newblock URL \url{http://dx.doi.org/10.1080/01621459.1977.10480613}.

\bibitem[Gerds(2023)]{pecpackage}
T.~A. Gerds.
\newblock \emph{Prediction Error Curves for Risk Prediction Models in Survival Analysis}, 2023.
\newblock URL \url{https://CRAN.R-project.org/package=pec}.
\newblock R package version 2023.04.12.

\bibitem[Gerds et~al.(2023)Gerds, Ohlendorff, Blanche, Mortensen, Wright, Tollenaar, Muschelli, Mogensen, and Ozenne]{riskRegressionpackage}
T.~A. Gerds, J.~S. Ohlendorff, P.~Blanche, R.~Mortensen, M.~Wright, N.~Tollenaar, J.~Muschelli, U.~B. Mogensen, and B.~Ozenne.
\newblock \emph{Risk Regression Models and Prediction Scores for Survival Analysis with Competing Risks}, 2023.
\newblock URL \url{https://CRAN.R-project.org/package=riskRegression}.
\newblock R package version 2023.12.21.

\bibitem[Graf et~al.(1999)Graf, Schmoor, Sauerbrei, and Schumacher]{Graf_1999}
E.~Graf, C.~Schmoor, W.~Sauerbrei, and M.~Schumacher.
\newblock Assessment and comparison of prognostic classification schemes for survival data.
\newblock \emph{Statistics in Medicine}, 18\penalty0 (17–18):\penalty0 2529–2545, September 1999.
\newblock ISSN 1097-0258.
\newblock \doi{10.1002/(sici)1097-0258(19990915/30)18:17/18<2529::aid-sim274>3.0.co;2-5}.
\newblock URL \url{http://dx.doi.org/10.1002/(SICI)1097-0258(19990915/30)18:17/18<2529::AID-SIM274>3.0.CO;2-5}.

\bibitem[Harrell et~al.(1996)Harrell, Lee, and Mark]{Harrell1996}
F.~E. Harrell, K.~L. Lee, and D.~B. Mark.
\newblock Multivariate prognostic models: Issues in developing models, evaluating assumptions and adequacy, and measuring and reducing errors.
\newblock \emph{Statistics in Medicine}, 15\penalty0 (4):\penalty0 361–387, February 1996.
\newblock ISSN 1097-0258.
\newblock \doi{10.1002/(sici)1097-0258(19960229)15:4<361::aid-sim168>3.0.co;2-4}.
\newblock URL \url{http://dx.doi.org/10.1002/(SICI)1097-0258(19960229)15:4<361::AID-SIM168>3.0.CO;2-4}.

\bibitem[He et~al.(2020)He, Fan, Wu, Xie, and Girshick]{he2020momentum}
K.~He, H.~Fan, Y.~Wu, S.~Xie, and R.~Girshick.
\newblock Momentum contrast for unsupervised visual representation learning.
\newblock In \emph{Proceedings of the IEEE/CVF conference on computer vision and pattern recognition}, pages 9729--9738, 2020.

\bibitem[Heagerty(2022{\natexlab{a}})]{risksetROCpackage}
P.~J. Heagerty.
\newblock \emph{Riskset ROC Curve Estimation from Censored Survival Data}, 2022{\natexlab{a}}.
\newblock URL \url{https://CRAN.R-project.org/package=risksetROC}.
\newblock R package version 1.0.4.1.

\bibitem[Heagerty(2022{\natexlab{b}})]{survivalROCpackage}
P.~J. Heagerty.
\newblock \emph{Time-Dependent ROC Curve Estimation from Censored Survival Data}, 2022{\natexlab{b}}.
\newblock URL \url{https://CRAN.R-project.org/package=survivalROC}.
\newblock R package version 1.0.3.1.

\bibitem[Heagerty and Zheng(2005)]{Heagerty2005}
P.~J. Heagerty and Y.~Zheng.
\newblock Survival model predictive accuracy and roc curves.
\newblock \emph{Biometrics}, 61\penalty0 (1):\penalty0 92–105, February 2005.
\newblock ISSN 1541-0420.
\newblock \doi{10.1111/j.0006-341x.2005.030814.x}.
\newblock URL \url{http://dx.doi.org/10.1111/j.0006-341x.2005.030814.x}.

\bibitem[Heagerty et~al.(2000)Heagerty, Lumley, and Pepe]{Heagerty2000}
P.~J. Heagerty, T.~Lumley, and M.~S. Pepe.
\newblock Time‐dependent roc curves for censored survival data and a diagnostic marker.
\newblock \emph{Biometrics}, 56\penalty0 (2):\penalty0 337–344, June 2000.
\newblock ISSN 1541-0420.
\newblock \doi{10.1111/j.0006-341x.2000.00337.x}.
\newblock URL \url{http://dx.doi.org/10.1111/j.0006-341x.2000.00337.x}.

\bibitem[Katzman et~al.(2018)Katzman, Shaham, Cloninger, Bates, Jiang, and Kluger]{katzman2018deepsurv}
J.~L. Katzman, U.~Shaham, A.~Cloninger, J.~Bates, T.~Jiang, and Y.~Kluger.
\newblock Deepsurv: personalized treatment recommender system using a cox proportional hazards deep neural network.
\newblock \emph{BMC medical research methodology}, 18\penalty0 (1):\penalty0 1--12, 2018.

\bibitem[Kvamme et~al.(2019)Kvamme, {{\O}}rnulf Borgan, and Scheel]{Kvamme2019pycox}
H.~Kvamme, {{\O}}rnulf Borgan, and I.~Scheel.
\newblock Time-to-event prediction with neural networks and cox regression.
\newblock \emph{Journal of Machine Learning Research}, 20\penalty0 (129):\penalty0 1--30, 2019.
\newblock URL \url{http://jmlr.org/papers/v20/18-424.html}.

\bibitem[Nagpal et~al.(2022)Nagpal, Potosnak, and Dubrawski]{nagpal2022auton}
C.~Nagpal, W.~Potosnak, and A.~Dubrawski.
\newblock auton-survival: An open-source package for regression, counterfactual estimation, evaluation and phenotyping with censored time-to-event data.
\newblock In \emph{Machine Learning for Healthcare Conference}, pages 585--608. PMLR, 2022.

\bibitem[Paszke et~al.(2019)Paszke, Gross, Massa, Lerer, Bradbury, Chanan, Killeen, Lin, Gimelshein, Antiga, et~al.]{paszke2019pytorch}
A.~Paszke, S.~Gross, F.~Massa, A.~Lerer, J.~Bradbury, G.~Chanan, T.~Killeen, Z.~Lin, N.~Gimelshein, L.~Antiga, et~al.
\newblock Pytorch: An imperative style, high-performance deep learning library.
\newblock \emph{Advances in neural information processing systems}, 32, 2019.

\bibitem[P{\"o}lsterl(2020)]{polsterl2020scikit}
S.~P{\"o}lsterl.
\newblock scikit-survival: A library for time-to-event analysis built on top of scikit-learn.
\newblock \emph{The Journal of Machine Learning Research}, 21\penalty0 (1):\penalty0 8747--8752, 2020.

\bibitem[Potapov et~al.(2023)Potapov, Adler, and Schmid]{survAUCpackage}
S.~Potapov, W.~Adler, and M.~Schmid.
\newblock \emph{Estimators of prediction accuracy for time-to-event data}, 2023.
\newblock URL \url{https://CRAN.R-project.org/package=survAUC}.
\newblock R package version 1.2-0.

\bibitem[Schr\"{o}der et~al.(2011)Schr\"{o}der, Culhane, Quackenbush, and Haibe-Kains]{survcomppackage}
M.~S. Schr\"{o}der, A.~C. Culhane, J.~Quackenbush, and B.~Haibe-Kains.
\newblock survcomp: an r/bioconductor package for performance assessment and comparison of survival models.
\newblock \emph{Bioinformatics}, 27\penalty0 (22):\penalty0 3206–3208, Sept. 2011.
\newblock ISSN 1367-4803.
\newblock \doi{10.1093/bioinformatics/btr511}.
\newblock URL \url{http://dx.doi.org/10.1093/bioinformatics/btr511}.

\bibitem[Therneau(2024)]{survivalpackage}
T.~M. Therneau.
\newblock \emph{A Package for Survival Analysis in R}, 2024.
\newblock URL \url{https://CRAN.R-project.org/package=survival}.
\newblock R package version 3.5-8.

\bibitem[Uno et~al.(2007)Uno, Cai, Tian, and Wei]{Uno2007}
H.~Uno, T.~Cai, L.~Tian, and L.~J. Wei.
\newblock Evaluating prediction rules fort-year survivors with censored regression models.
\newblock \emph{Journal of the American Statistical Association}, 102\penalty0 (478):\penalty0 527–537, June 2007.
\newblock ISSN 1537-274X.
\newblock \doi{10.1198/016214507000000149}.
\newblock URL \url{http://dx.doi.org/10.1198/016214507000000149}.

\bibitem[Uno et~al.(2011)Uno, Cai, Pencina, D’Agostino, and Wei]{Uno_2011}
H.~Uno, T.~Cai, M.~J. Pencina, R.~B. D’Agostino, and L.~J. Wei.
\newblock On the c‐statistics for evaluating overall adequacy of risk prediction procedures with censored survival data.
\newblock \emph{Statistics in Medicine}, 30\penalty0 (10):\penalty0 1105–1117, January 2011.
\newblock ISSN 1097-0258.
\newblock \doi{10.1002/sim.4154}.
\newblock URL \url{http://dx.doi.org/10.1002/sim.4154}.

\bibitem[Zhou et~al.(2022)Zhou, Cheng, Wang, Zou, and Wang]{SurvMetricspackage}
H.~Zhou, X.~Cheng, S.~Wang, Y.~Zou, and H.~Wang.
\newblock \emph{Predictive Evaluation Metrics in Survival Analysis}, 2022.
\newblock URL \url{https://CRAN.R-project.org/package=SurvMetrics}.
\newblock R package version 0.5.0.

\end{thebibliography}
